\documentclass{article}

\usepackage[nonatbib,final]{nips_2016}

\usepackage[T1]{fontenc}    % use 8-bit T1 fonts
\usepackage{hyperref}       % hyperlinks
\usepackage{url}            % simple URL typesetting
\usepackage{booktabs}       % professional-quality tables
\usepackage{amsfonts}       % blackboard math symbols
\usepackage{nicefrac}       % compact symbols for 1/2, etc.
\usepackage{microtype}      % microtypography

\usepackage{amsmath}
\usepackage{amssymb}
\usepackage[pdftex]{graphicx}
\usepackage{multirow}
\usepackage{graphicx}
\usepackage{epstopdf}
\usepackage{epsfig}
\usepackage{caption}
\usepackage{subcaption}
\usepackage{amsthm}
\usepackage{multirow}

\usepackage{algorithmic}
\usepackage[lined,ruled,commentsnumbered]{algorithm2e}

\newcommand{\boldfrac}[2]{\genfrac{}{}{1pt}{}{#1}{#2}}
\newcommand\Mark[1]{\textsuperscript#1}
\usepackage{textcomp}

\pdfoutput=1

\title{Efficient Convolutional Auto-Encoding via Random Convexification and Frequency-Domain Minimization}

\author{
  Meshia C. Oveneke\Mark{1}, Mitchel Aliosha-Perez\Mark{1}, Yong Zhao\Mark{1}\Mark{2}, Dongmei Jiang\Mark{2}, Hichem Sahli\Mark{1}\Mark{3} \\ \\
  \Mark{1}VUB-NPU Joint AVSP Research Lab \\
  Vrije Universiteit Brussel (VUB) \\
  Deptartment of Electronics \& Informatics (ETRO) \\
  Pleinlaan 2, Brussels, Belgium \\
  \texttt{\{mcovenek,maperezg,yzhao,hsahli\}@etrovub.be} \\ 
  \Mark{2}VUB-NPU Joint AVSP Research Lab \\
  Northwestern Polytechnical University (NPU)\\
  Shaanxi Key Lab on Speech and Image Information Processing \\
  Youyo Xilu 127, Xi'an 710072, China\\
  \texttt{jiangdm@nwpu.edu.cn} \\ 
  \Mark{3}Interuniversity Microelectronics Centre (IMEC)\\
  Kapeldreef 75, 3001 Heverlee, Belgium
}

\begin{document}

\maketitle

\begin{abstract}
	The omnipresence of deep learning architectures such as deep \textit{convolutional neural networks} (CNN)s is fueled by the synergistic combination of ever-increasing labeled datasets and specialized hardware.
	Despite the indisputable success, the reliance on huge amounts of labeled data and specialized hardware can be a limiting factor when approaching new applications.
	To help alleviating these limitations, we propose an efficient learning strategy for layer-wise unsupervised training of deep CNNs on conventional hardware in acceptable time.
	Our proposed strategy consists of randomly convexifying the \textit{reconstruction contractive auto-encoding} (RCAE) learning objective and solving the resulting large-scale convex minimization problem in the frequency domain via \textit{coordinate descent} (CD).
	The main advantages of our proposed learning strategy are: (1) single tunable optimization parameter; (2) fast and guaranteed convergence; (3) possibilities for full parallelization.
	Numerical experiments show that our proposed learning strategy scales (in the worst case) linearly with image size, number of filters and filter size. 
\end{abstract}

\section{Introduction}\label{sec_Introduction}

At the heart of the recent success of deep \textit{convolutional neural networks} (CNN)s in several application domains such as computer vision, speech recognition and natural language processing, is the synergistic combination of ever-increasing \textit{labeled datasets} and \textit{specialized hardware} in the form of graphical processing units (GPU)s \cite{raina2009}, field programmable gate arrays (FPGA)s \cite{lacey2016} or computer clusters \cite{wu2015}.
Despite this indisputable success, relying on huge amounts of labeled data and specialized hardware can be a limiting factor when approaching new applications.
Furthermore, for applying deep learning techniques locally on platforms with limited resources, one cannot rely on huge amounts of labeled data and specialized hardware. 
We therefore argue that there is a growing need for resource-efficient \textit{unsupervised learning strategies} capable of training deep CNNs on conventional hardware in acceptable time. 
The main purpose of unsupervised learning is to leverage the wealth of unlabeled data for disentangling the causal (generative) factors \cite{bengio2013,higgins2016}.
In the context of supervised pattern classification, empirical evidence shows that: unsupervised pre-training (generative) helps in disentangling the class manifolds in the lower layers of a deep CNNs, the upper layers are better disentangled when they are subject to supervised training (discriminative) and a combination of the two boosts the overall performance when the ratio of unlabeled to labeled samples is high \cite{brahma2016,erhan2010,paine2014}.

In this work, we propose to train deep CNNs in a greedy layer-wise manner, using the \textit{reconstruction contractive auto-encoding} (RCAE) learning objective \cite{alain2014}.
The RCAE learning objective has been proven (theoretically and empirically) to capture the local shape of the data-generating distribution \cite{bengio2013,alain2014}, hence capturing the manifold structure of the data.
Meanwhile, minimizing the RCAE learning objective involves solving a non-convex minimization problem, often addressed using \textit{stochastic gradient descent} (SGD) methods \cite{ngiam2011}.
Despite their empirical success when applied to deep CNNs, their key disadvantage is the computationally expensive manual tuning of optimization parameters such as learning rates and convergence criteria. 
Moreover, their inherently sequential nature makes them very difficult to parallelize using GPUs or distribute them using computer clusters \cite{ngiam2011}. 
To overcome the above-mentioned difficulties, we propose to convexify the RCAE learning objective.
To this end, inspired by recent work such as \cite{rahimi2009,huang2015,liu2015}, we propose to adopt a \textit{random convexification} strategy by fixing the (non-linear) encoding parameters and only learning the (linear) decoding parameters.
By further transforming the randomly convexified RCAE objective into the frequency domain using the \textit{discrete Fourier transform} (DFT), we obtain a learning objective which we propose to minimize using \textit{coordinate descent} (CD) \cite{nesterov2012,wright2015}.
The main advantages of our proposed learning strategy are: (1) single tunable optimization parameter; (2) fast and guaranteed convergence; (3) possibilities for full parallelization.

\section{Efficient Convolutional Auto-Encoding} \label{sec_CAE}
The main motivation of this work is efficient unsupervised training of deep CNNs.
To this end, we adopt a greedy layer-wise learning strategy and consider deep CNNs as a stack of single-layer CNNs with $K$ filters and input space $\mathcal{X} \subset \mathbb{R}^{d \times d \times C}$, i.e. the space of $C$-channel $d \times d$ images \footnote{Although we consider $d \times d$ images, the remaining calculations also hold for rectangular images.} $\mathbf{x} = \left(\mathbf{x}^{(1)},\ldots,\mathbf{x}^{(C)}\right)$.
We further propose to train each single-layer CNN using the \textit{reconstruction contractive auto-encoding} (RCAE) learning objective \cite{alain2014}.
The RCAE learning objective has been proven to capture the high-density regions of the (layer-wise) data-generating distribution for any twice-differential reconstruction (encoding-decoding) function and regularization parameter approaching $0$ asymptotically.
To this end, we consider the following parametric reconstruction function:
\begin{equation} \label{eq_CAE}
\begin{split}
\mathbf{r}(\mathbf{x};\boldsymbol\theta) & \triangleq \underbrace{\sum_{k=1}^K \mathbf{w}^{(k)}}_{\text{decoding}} \ast \underbrace{\vphantom{\sum_{k=1}^K} \mathbf{g}(\sum_{c=1}^C \mathbf{a}^{(k)}\ast\mathbf{x}^{(c)} + \mathbf{b}^{(k)})}_{\text{encoding}} \\
& = \sum_{k=1}^K \mathbf{w}^{(k)} \ast \mathbf{h}^{(k)}
\end{split}
\end{equation}
with $\mathbf{g}(\cdot)$ denoting the entry-wise application of a twice-differentiable activation function $g:\mathbb{R} \to \mathbb{R}$. 
The model parameters $\boldsymbol\theta $ consist of $w \times w$ encoding filters $\{\mathbf{a}^{(k)}\}_{k=1}^K$, $h \times h$ encoding biases $\{\mathbf{b}^{(k)}\}_{k=1}^K$ and $w \times w$ decoding filters $\{\mathbf{w}^{(k)}\}_{k=1}^K$. 
We consider a valid convolution for the encoding function, yielding a dimension $h = d-w+1$, and a full convolution for the decoding function.

The core idea behind our proposed layer-wise learning strategy is to sample each entry of the (non-linear) encoding parameters $\{\mathbf{a}^{(k)}\}_{k=1}^K$ and $\{\mathbf{b}^{(k)}\}_{k=1}^K$  i.i.d. from pre-determined density functions $p(a)$ and $p(b)$ respectively, and keeping them fixed while learning the (linear) decoding parameters $\{\mathbf{w}^{(k)}\}_{k=1}^K$ in the frequency domain.
To this end, we define the complex-valued decoding parameters associated to $\{\mathbf{w}^{(k)}\}_{k=1}^K$ as $\{\mathbf{W}^{(k)}\}_{k=1}^K$, with $\mathbf{W}^{(k)} = \mathcal{F}\{\mathbf{w}^{(k)}\} \in \mathbb{C}^{d \times d}$ being the \textit{discrete Fourier transform} (DFT). 
Given a set of $N$ training images $\mathcal{D}_N = \{\mathbf{x}_n\}_{n=1}^N$ sampled i.i.d. from the data-generating distribution $p(\mathbf{x})$, we define the following frequency-domain RCAE objective:
\begin{equation} \label{eq_LossFourier}
\begin{split}
\mathcal{L}_{\text{RCAE}} & = \int_{\mathcal{X}} p(\mathbf{x})\left[
\left\Vert \mathbf{r}(\mathbf{x};\boldsymbol\theta)-\mathbf{x} \right\Vert_F^2 + 
\lambda \left\Vert \frac{\partial \mathbf{r}(\mathbf{x};\boldsymbol\theta)}{\partial \mathbf{x}} \right\Vert_F^2 
\right]d\mathbf{x}  \\
& \approx \frac{1}{N} \sum_{n=1}^N \left[
\left\Vert \mathbf{r}(\mathbf{x}_n;\boldsymbol\theta)-\mathbf{x}_n \right\Vert_F^2 + 
\lambda \left\Vert \left. \frac{\partial \mathbf{r}(\mathbf{x};\boldsymbol\theta)}{\partial \mathbf{x}} \right\vert_{\mathbf{x}=\mathbf{x}_n} \right\Vert_F^2 \right]\\
& \propto \frac{1}{N} \sum_{n=1}^N \left[
\left\Vert \mathcal{F}\{\mathbf{r}(\mathbf{x}_n;\boldsymbol\theta)-\mathbf{x}_n\} \right\Vert_F^2 + 
\lambda \left\Vert \mathcal{F}\left\lbrace \left. \frac{\partial \mathbf{r}(\mathbf{x};\boldsymbol\theta)}{\partial \mathbf{x}} \right\vert_{\mathbf{x}=\mathbf{x}_n} \right\rbrace \right\Vert_F^2 \right]\\
& = \frac{1}{N} \sum_{n=1}^N \left[
\left\Vert \sum_{k=1}^K \mathbf{W}^{(k)} \odot \mathbf{H}^{(k)}_n - \mathbf{X}_n \right\Vert_F^2 +  \lambda \left\Vert \sum_{k=1}^K \mathbf{W}^{(k)} \odot \mathbf{D}^{(k)}_n \right\Vert_F^2 \right]\\
\end{split}
\end{equation} 
where $\Vert \bullet \Vert_F$ is the Frobenius norm, $\odot$ denotes the Hadamard (entry-wise) product \cite{golub2012}, $\mathbf{H}^{(k)}_n = \mathcal{F}\{\mathbf{h}^{(k)}_n\}$, $\mathbf{X}_n = \sum_{c=1}^C\mathcal{F}\{\mathbf{x}^{(c)}_n\}$, $\mathbf{D}^{(k)}_n = \mathbf{G}^{(k)}_n \odot \mathbf{A}^{(k)}$ with $\mathbf{G}^{(k)}_n = \mathcal{F}\left\lbrace \left. \frac{\partial g(\mathbf{v})}{\partial \mathbf{v}} \right\vert_{\mathbf{v}=\mathbf{v}^{(k)}_n}\right\rbrace$ and $\mathbf{v}^{(k)}_n = \sum_{c=1}^C \mathbf{a}^{(k)}\ast\mathbf{x}^{(c)}_n + \mathbf{b}^{(k)}$.
We used Parseval's theorem in conjunction with the convolution theorem \cite{kammler2007} for transforming the randomly convexified RCAE objective from spatial to frequency domain.
As a direct consequence, minimizing the transformed RCAE objective (\ref{eq_LossFourier}) reduces to solving $d^2$ independent $K$-dimensional regularized linear least-squares problems, which offers full parallelization possibilities.
We propose to solve each of the independent $K$-dimensional problems using \textit{coordinate descent} (CD) \cite{wright2015}, which has recently witnessed a resurgence of interest in large-scale optimization problems due to its simplicity and fast convergence \cite{nesterov2012}.
In the context of RCAE, the CD method consists of iteratively minimizing (\ref{eq_LossFourier}) along the $k$-th coordinate (filter), while keeping the remaining $K-1$ coordinates (filters) fixed, yielding the following strategy for updating the filters in the frequency domain:
\begin{equation} \label{eq_CCD}
\hat{\mathbf{W}}^{(k)} = \boldfrac{\bar{\mathbf{H}}_N^{(k)}\odot\mathbf{X}_N - \sum_{i \neq k}\hat{\mathbf{W}}^{(i)}\odot\left(\mathbf{H}_N^{(i)}\odot\bar{\mathbf{H}}_N^{(k)}  +  \lambda\mathbf{D}_N^{(i)}\odot\bar{\mathbf{D}}_N^{(k)}\right)}{
\mathbf{H}_N^{(k)}\odot\bar{\mathbf{H}}_N^{(k)} + \lambda\mathbf{D}_N^{(k)}\odot\bar{\mathbf{D}}_N^{(k)}}
\end{equation}
where the bold fraction sign ($\boldfrac{\cdot}{\cdot}$) denotes the Hadamard (entry-wise) division and $\mathbf{H}^{(k)}_N = \sum_{n=1}^N\mathbf{H}^{(k)}_n$, $\mathbf{X}_N = \sum_{n=1}^N\mathbf{X}_n$ and $\mathbf{D}^{(k)}_N = \sum_{n=1}^N\mathbf{D}^{(k)}_n$.
The main advantages of the derived filter update strategy (\ref{eq_CCD})  are: single tunable optimization parameter $\lambda$, fast and guaranteed convergence of $\hat{\mathbf{W}}^{(k)}$; possibilities for parallelization offered by the frequency-domain transformation of $\mathcal{L}_{\text{RCAE}}$.
It is important to note that the incremental data acquisition involved in $\mathbf{H}^{(k)}_N$, $\mathbf{X}_N$ and $\mathbf{D}^{(k)}_N$ can also be computed in parallel.
Once the decoding filters are learned in the frequency domain, they are transformed back to the spatial domain using the inverse DFT:
\begin{equation}
\hat{\mathbf{w}}^{(k)} = \mathcal{F}^{-1}\{\hat{\mathbf{W}}^{(k)}\}
\end{equation} 
At inference stage, we use the transpose of the learned decoding filters $\hat{\mathbf{w}}^{(k)}$ for computing $k$ feature maps as follows: $\hat{\mathbf{h}}^{(k)} \triangleq \mathbf{g}(\sum_{c=1}^C\hat{\mathbf{w}}^{(k)T} \ast \mathbf{x}^{(c)})$.

\section{Numerical Experiments} \label{sec_Experiments}

\begin{figure*}[!t]
	\centering
	\begin{tabular}{lcr}
		\includegraphics[scale=0.14]{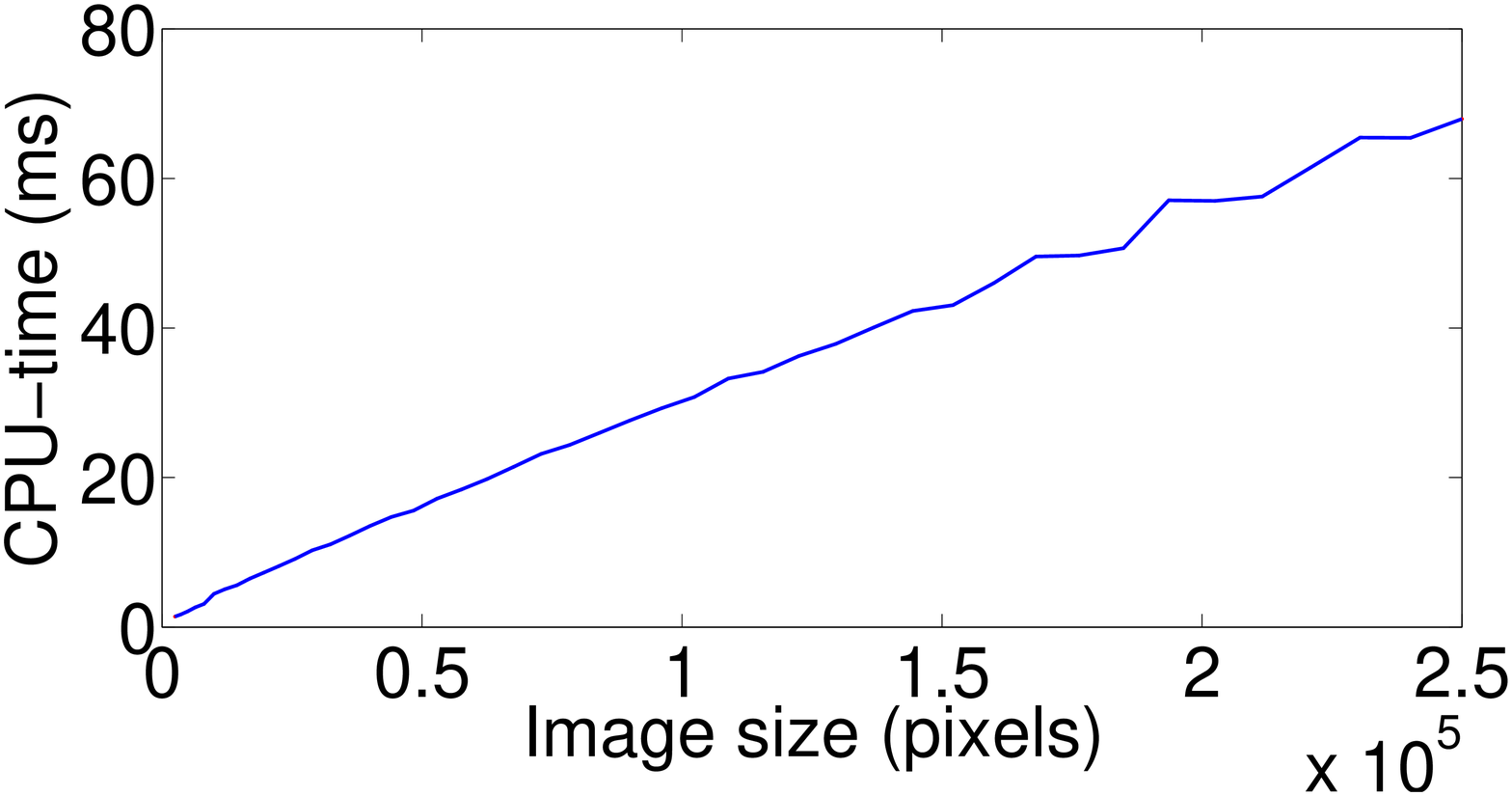}
		\includegraphics[scale=0.14]{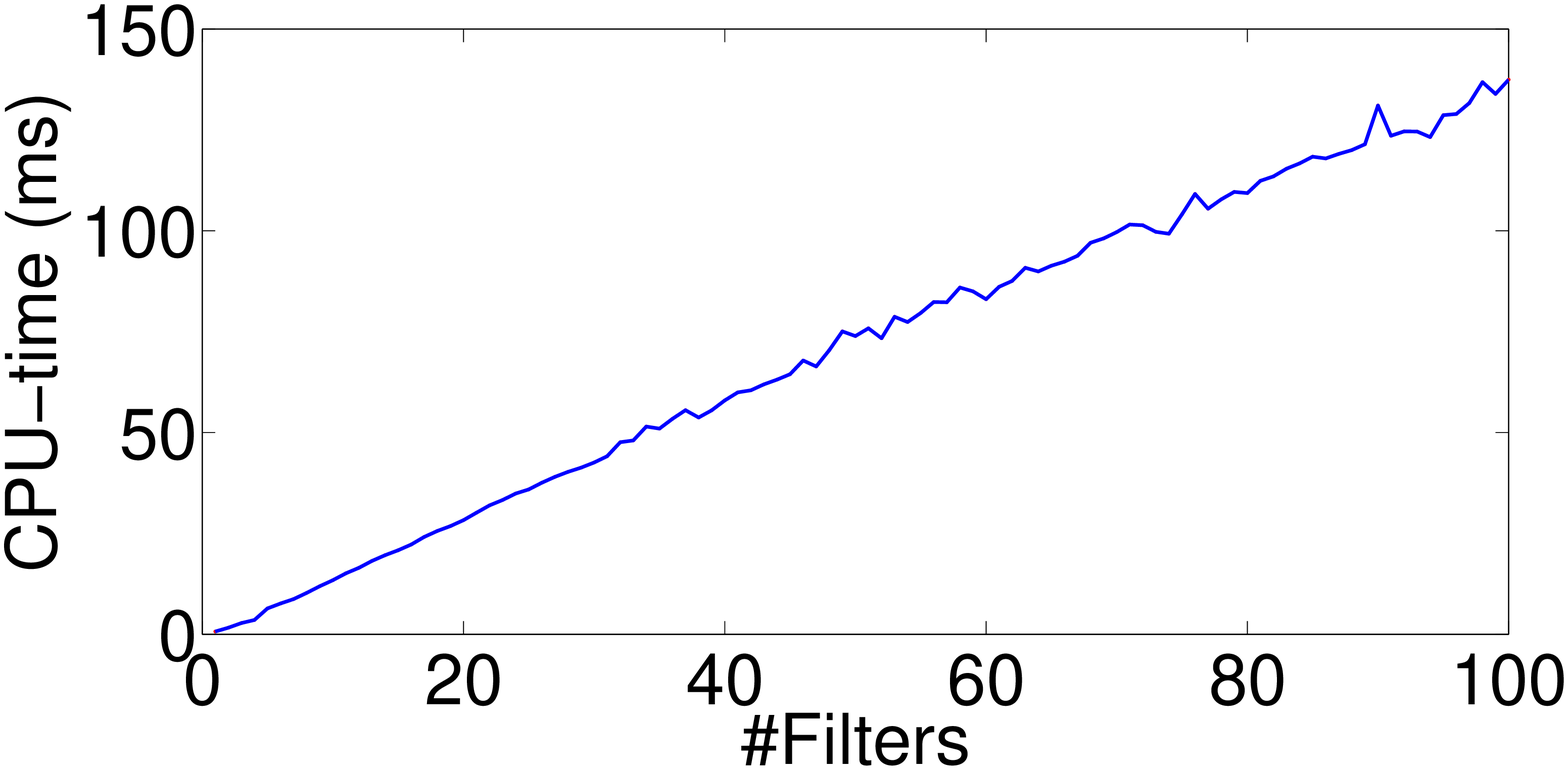}
		\includegraphics[scale=0.14]{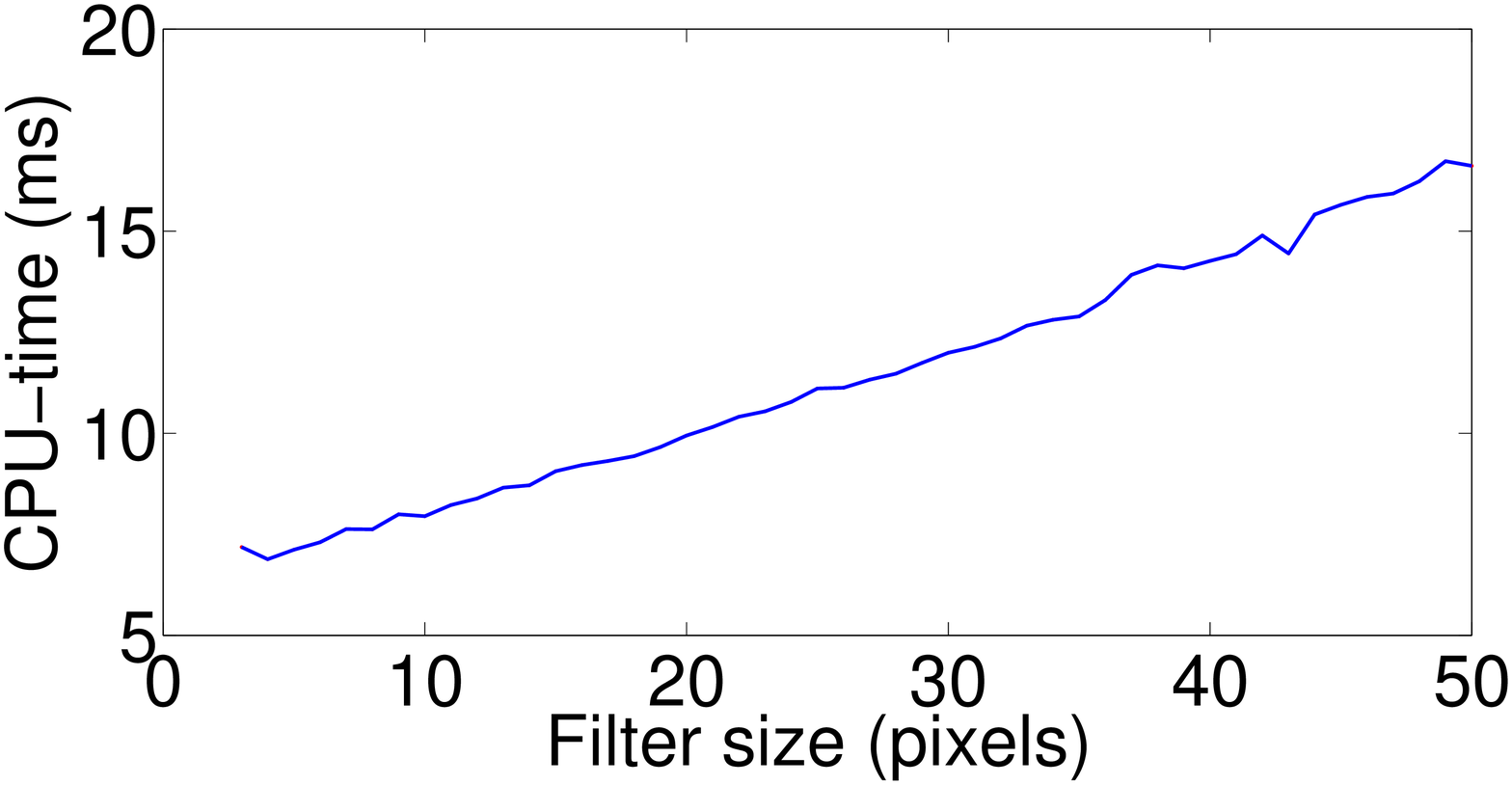}
	\end{tabular}
	\caption{Time-complexity of CD-based RCAE minimization on Caltech-256 dataset.
		\textit{Left}: CPU-time (in milliseconds) in function of image size (in pixels).
		\textit{Middle}: CPU-time (in milliseconds) in function of number of filters.
		\textit{Right}: CPU-time (in milliseconds) in function of filter size (in pixels).}
	\label{fig_Scalability}
\end{figure*}

\begin{figure*}[!t]
	\begin{tabular}{cc}
		\includegraphics[scale=0.17]{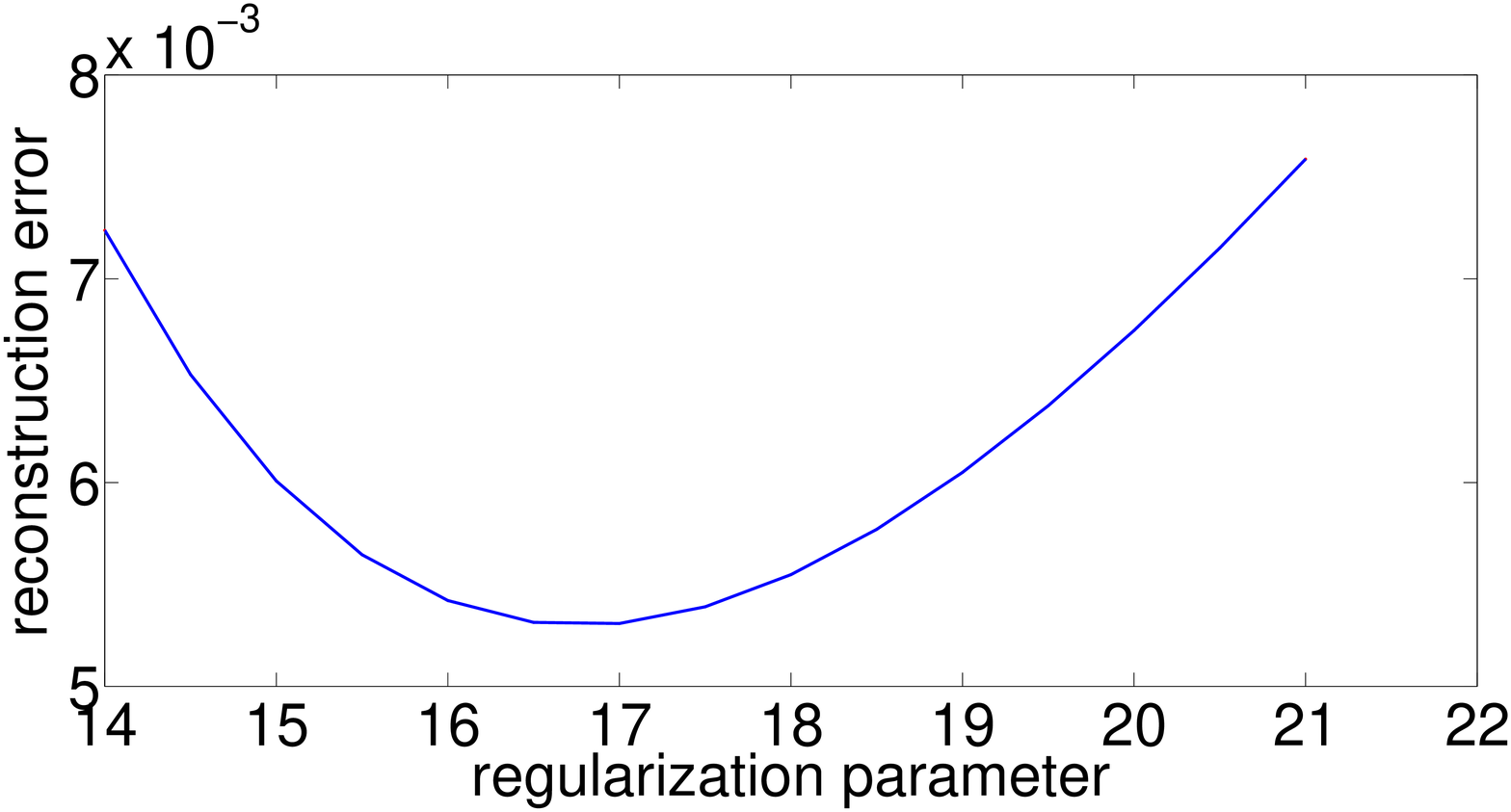}
		\includegraphics[scale=0.17]{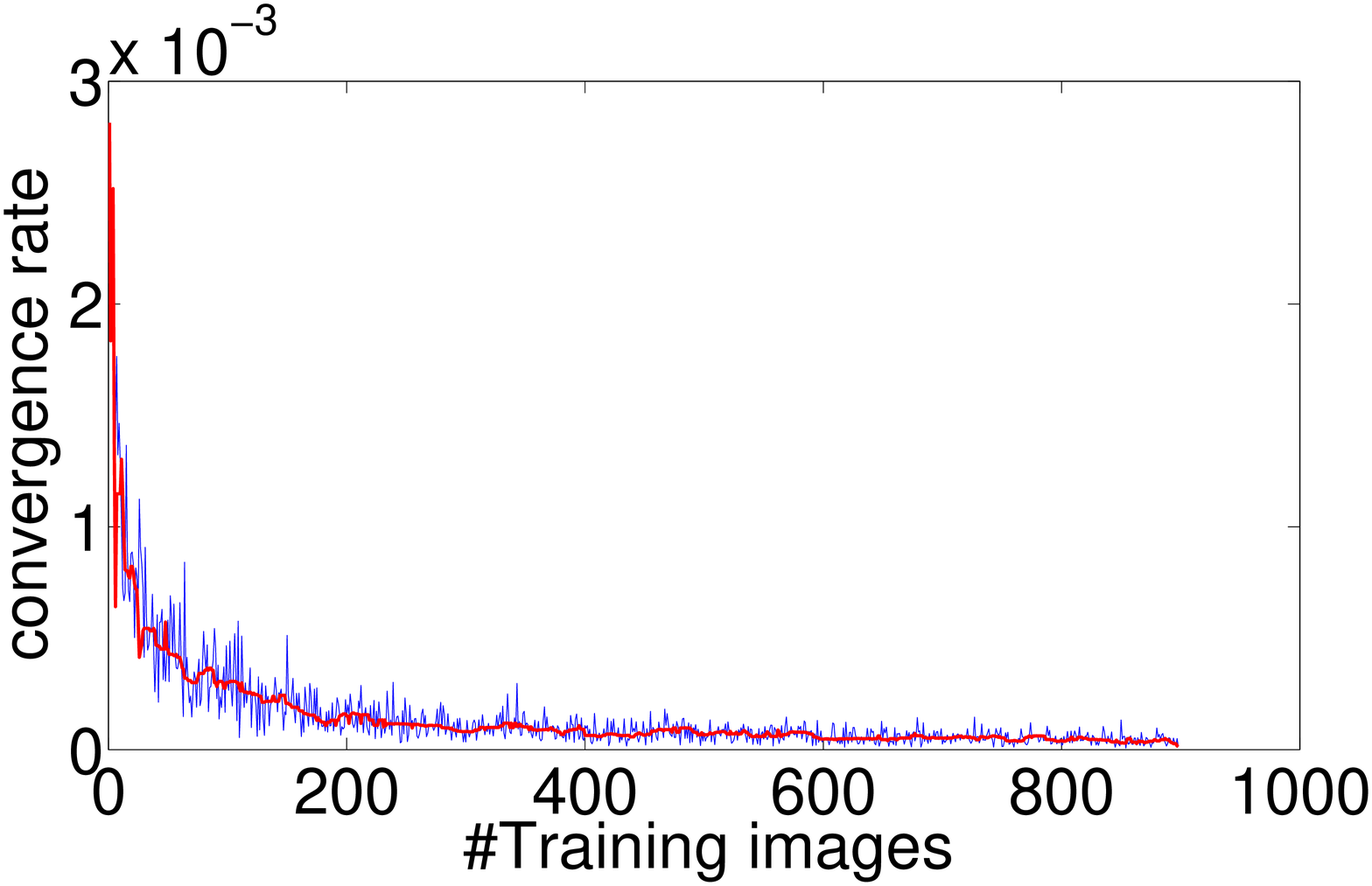}
	\end{tabular}
	\caption{Reconstruction error and convergence rate of CD-based RCAE minimization on Caltech-256 dataset.
		\textit{Left}: Reconstruction error in function of regularization parameter $\lambda$.
		\textit{Right}: Convergence rate of the learned decoding filters in function of number of training images $N$.
		The convergence rate is computed using the average squared difference between consecutive filter updates.
		The original curve (blue) is noisy due to the average involved in the computation of the convergence rate. 
		A smoothed curve (red) is therefore plotted on top, in order to highlight the general trend.}
	\label{fig_ReconstructionConvergence}
\end{figure*}
\begin{figure}[!ht]
	\centering
	\minipage{0.35\textwidth}
	\includegraphics[width=\linewidth]{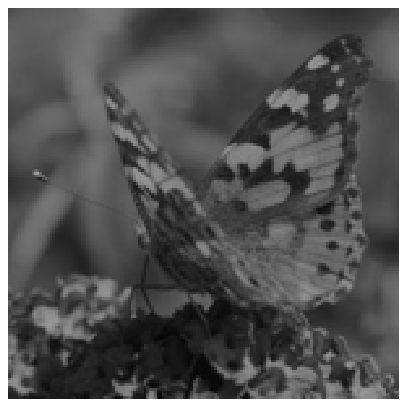}
	\endminipage
	\minipage{0.35\textwidth}%
	\includegraphics[width=\linewidth]{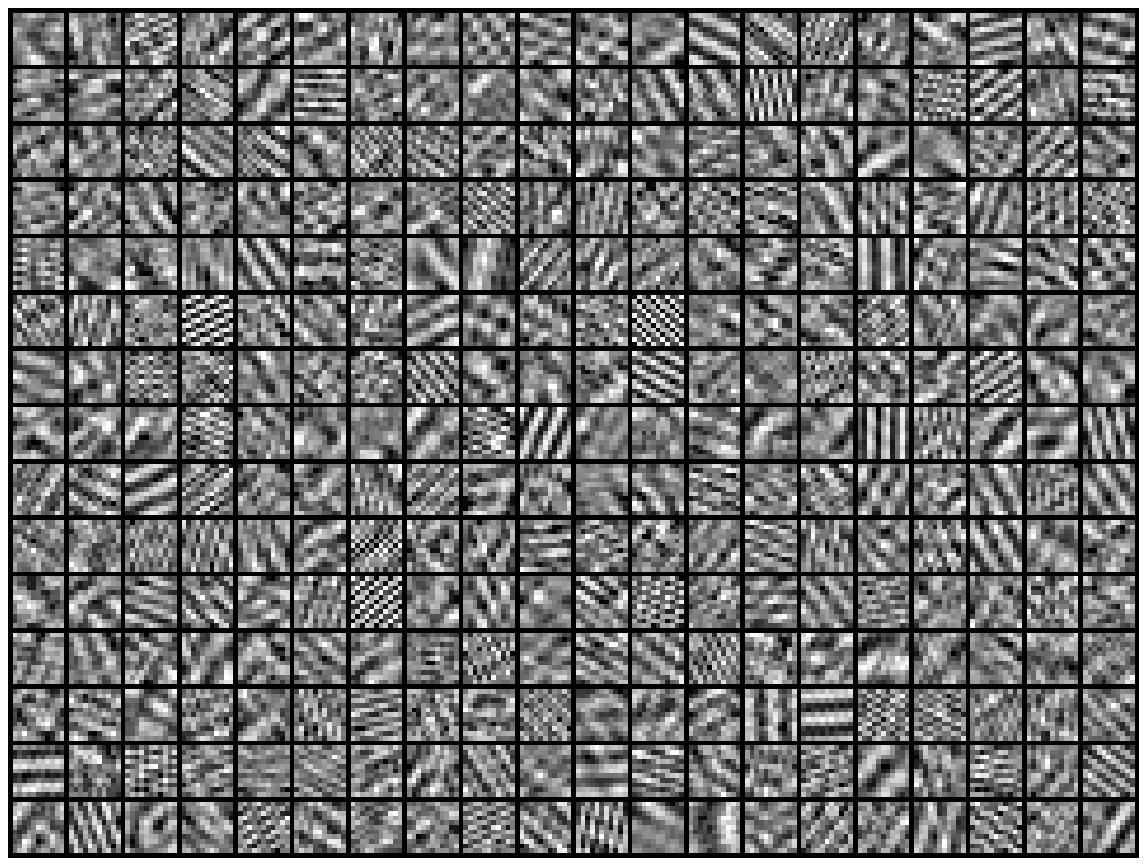}
	\endminipage
	\minipage{0.35\textwidth}%
	\includegraphics[width=\linewidth]{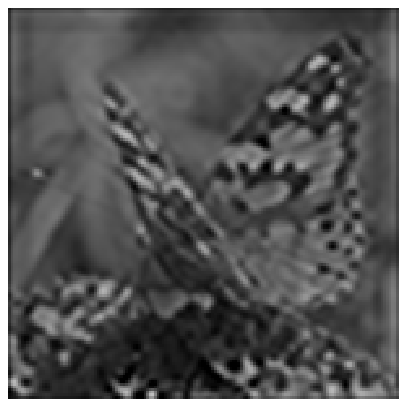}
	\endminipage
	\caption{Visualization of learned decoding filters and reconstruction result on Caltech-256 dataset (after training with $400$ images).
	\textit{Left}: Previously unseen $244 \times 244$ testing image.
	\textit{Middle}: $300$ learned decoding filters.
	\textit{Right}: Reconstruction result on previously unseen $244 \times 244$ testing image (left) using $300$ learned decoding filters (middle).}
	\label{fig_LearnedFilters_Caltech256}
\end{figure}

As a sanity check for the overall computational efficiency of our proposed learning strategy, we've implemented the CD-based frequency-domain RCAE minimization in MATLAB R2014a.
%\footnote{The source code is available at ~\url{http://www.etrovub.be/RESEARCH/AVSP/Downloads/AVSP_CURL/}}
We've used the built-in \textit{fast Fourier transform} (FFT) for transforming the RCAE objective into the frequency domain.
The hyperbolic tangent was used as activation function.
The encoding filters and biases were randomly fixed by sampling their entries independently from a zero-mean normal distribution with standard deviations $0.1$ and $0.01$ respectively.
For learning the decoding filters, we've implemented a fairly naive version of frequency-domain CD (\ref{eq_CCD}) by initializing each complex-valued filter as $\hat{\mathbf{W}}^{(k)} = \mathbf{0} + \mathfrak{i}\mathbf{0}$ and only performing a single CD cycle trough the filter coordinates $k \in [1,K]$.
For all the experiments, we used of the Caltech-256 Object Category dataset \cite{griffin2007} and whitened the images.

As first experiment, we've measured the computational time-complexity in terms of CPU-time on an Intel\textregistered~Core\texttrademark~i7-2600 CPU @ 3.40 GHz machine. 
Figure (\ref{fig_Scalability}) shows that our proposed method has a (worst-case) linear time-complexity w.r.t. image size, number of filters and filter size.
Knowing that our naive implementation does not involve any particular form of parallelism, linear time is the best possible complexity we can achieve in situations where the algorithm has to read its entire input sequentially.
In a second experiment, we study the influence of the regularization parameter $\lambda$ on the reconstruction error and analyze the convergence rate of the learned decoding filters in function of the number training samples $N$.
For robust estimation, the reconstruction error was averaged over a batch of $500$ images that were not used during training.
Figure (\ref{fig_ReconstructionConvergence}), left, illustrates that the reconstruction error reaches a (global) minimum when the regularization parameter approaches $\lambda=16.5$.
Figure (\ref{fig_ReconstructionConvergence}), right, illustrates that after roughly $400$ training images, the learned decoding filters $\{\hat{\mathbf{w}}^{(k)}\}_{k=1}^K$  already settle.
This clearly highlights the advantages of random convexification and frequency-domain minimization using CD. 
Figure (\ref{fig_LearnedFilters_Caltech256}) depicts the decoding filters and reconstruction result on a previously unseen testing image, obtained after CD-based minimization of the RCAE objective (\ref{eq_LossFourier}) using only $400$ training images.

\section{Conclusions} \label{sec_Conclusion}
We've proposed an efficient learning strategy for layer-wise unsupervised training of deep CNNs.
The main contributions of our proposed learning strategy are random convexification and frequency-domain transformation of the \textit{reconstruction contractive auto-encoding} (RCAE) objective, which yields a relatively easy-to-solve large-scale regularized linear least-squares problem.
We've proposed to solve this problem using \textit{coordinate descent} (CD), with as main advantages: (1) single tunable optimization parameter; (2) fast and guaranteed convergence; (3) possibilities for full parallelization.
Numerical experiments show that, with a fairly naive implementation, our proposed learning strategy scales (in the worst case) linearly with image size, number of filters and filter size. 
We also observe that, using relatively few training images, the learned filters already settle and yield decent reconstruction results on unseen testing images.
We believe that the inherently parallel nature of our proposed learning strategy offers very interesting possibilities for further increasing the computational efficiency using more sophisticated implementation strategies.

\subsubsection*{Acknowledgments}
This work is supported by the Agency for Innovation by Science and Technology in Flanders (IWT) -- PhD grant nr. 131814, the VUB Interdisciplinary Research Program through the EMO-App project and the National Natural Science Foundation of China (grant 61273265). 

\bibliographystyle{plain}
\bibliography{References}

\end{document}